\pdfoutput=1
\documentclass{article}
\usepackage[colorlinks = true,
            linkcolor = blue,
            urlcolor  = blue,
            citecolor = blue,
            anchorcolor = blue]{hyperref}
\usepackage{natbib}
\usepackage{graphicx}
\usepackage{etoolbox}

\title{Seven Myths in Machine Learning Research}
\author{Oscar Chang, Hod Lipson}
\date{February 2019}

\begin{document}

\maketitle
\begin{abstract}
We present seven myths commonly believed to be true in machine learning research, circa Feb 2019. This is an archival copy of the blog post at \href{https://crazyoscarchang.github.io/2019/02/16/seven-myths-in-machine-learning-research/}{https://crazyoscarchang.github.io/2019/02/16/seven-myths-in-machine-learning-research/}\\
\hyperref[sec:myth-1]{Myth 1:} TensorFlow is a Tensor manipulation library\\
\hyperref[sec:myth-2]{Myth 2:} Image datasets are representative of real images found in the wild\\
\hyperref[sec:myth-3]{Myth 3:} Machine Learning researchers do not use the test set for validation\\
\hyperref[sec:myth-4]{Myth 4:} Every datapoint is used in training a neural network\\
\hyperref[sec:myth-5]{Myth 5:} We need (batch) normalization to train very deep residual networks\\
\hyperref[sec:myth-6]{Myth 6:} Attention $>$ Convolution\\
\hyperref[sec:myth-7]{Myth 7:} Saliency maps are robust ways to interpret neural networks
\end{abstract}

\section*{Myth 1: TensorFlow is a Tensor manipulation library}
\label{sec:myth-1}
It is actually a \emph{Matrix} manipulation library, and this difference is significant.

In \citet{laue2018computing}, the authors demonstrate that their automatic differentiation library based on actual Tensor Calculus has significantly more compact expression trees. This is because Tensor Calculus uses index notation, which results in treating both the forward mode and the reverse mode in the same manner.

By contrast, Matrix Calculus hides the indices for notational convenience, and this often results in overly complicated automatic differentiation expression trees.

Consider the matrix multiplication $ C = AB $. We have $ \dot{C} = \dot{A}B + A\dot{B} $ for the forward mode and $ \bar{A} = \bar{C}B^T, \bar{B} = A^T \bar{C} $ for the reverse mode. To perform the multiplications correctly, we have to be careful about the order of multiplication and the use of transposes. Notationally, this is a point of confusion for the machine learning practitioner, but computationally, this is an overhead for the program.

Here's another example, which is decidedly less trivial: $ c = det(A) $. We have $ \dot{c} = tr(inv(A)\dot{A}) $ for the forward mode, and $ \bar{A} = \bar{c}c inv(A)^T $ for the reverse mode. In this case, it is clearly not possible to use the same expression tree for both modes, given that they are composed of different operations.

In general, the way TensorFlow and other libraries (e.g. Mathematica, Maple, Sage, SimPy, ADOL-C, TAPENADE, TensorFlow, Theano, PyTorch, HIPS autograd) implement automatic differentiation results in different and inefficient expression trees for the forward and reverse mode. Tensor calculus conveniently avoids these problems by having commutativity in multiplication as a result of its index notation. (Please read the actual paper to learn more about how this works.)

The authors tested their method of doing reverse-mode automatic differentiation, aka backpropagation, on three different problems and measured the amount of time it took to compute the Hessians.

\includegraphics[width=\textwidth]{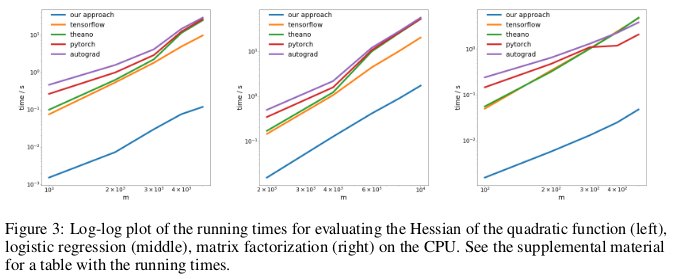}

The first problem involves optimizing a quadratic function like $ x^T A x$. The second problem solves for logistic regression, while the third problem solves for matrix factorization.

On the CPU, their method was faster than popular automatic differentiation libraries like TensorFlow, Theano, PyTorch, and HIPS autograd by \emph{two} orders of magnitude.

\includegraphics[width=\textwidth]{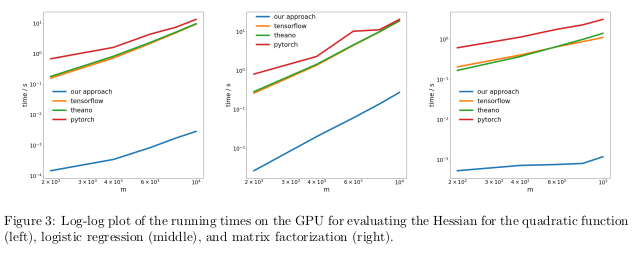}

On the GPU, they observed an even greater speedup, outperforming these libraries by a factor of \emph{three} orders of magnitude.

\subsection*{Implication:}
Computing derivatives for quadratic or higher functions with current deep learning libraries is more expensive than it needs to be. This includes computing general fourth order tensors like the Hessian (e.g. in MAML and second-order Newton optimization). Fortunately, quadratic functions are not common in “deep learning.” But they are common in “classical” machine learning - dual of an SVM, least squares regression, LASSO, Gaussian Processes, etc.

\section*{Myth 2: Image datasets are representative of real images found in the wild}
\label{sec:myth-2}
We like to think that neural networks are now better than humans at the task of object recognition. This is not true. They might outperform humans on select image datasets like ImageNet, but given actual images found in the wild, they are most definitely not going to be better than a regular adult human at recognizing objects. This is because images found in current image datasets are not actually drawn from the same distribution as the set of all possible images naturally occurring in the wild.

In an old paper \citet{torralba2011unbiased}, the authors proposed to examine dataset bias in twelve popular image datasets by observing if it is possible to train a classifier to identify the dataset a given image is selected from.

\includegraphics[width=\textwidth]{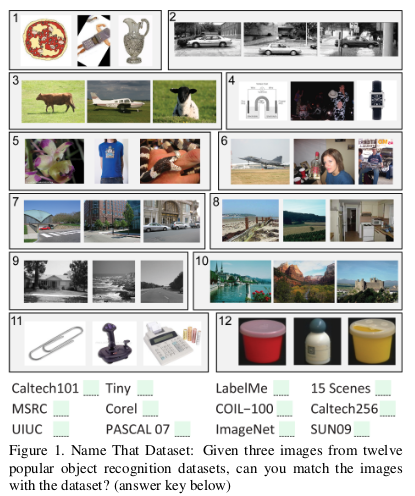}

The chance of getting it right by random is $ \frac{1}{12} \approx 8\% $, while their lab members performed at $>75\%$.

\includegraphics[width=\textwidth]{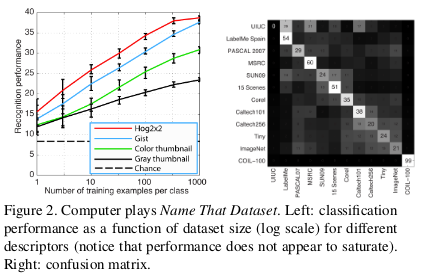}

They trained an SVM on HOG features, and found that their classifier performed at $ 39\% $, way above chance. If the same experiment was repeated today with a state of the art CNN, we will probably see a further increase in classifier performance.

If image datasets are truly representative of real images found in the wild, we ought to not be able to distinguish which dataset a given image originates from.

\includegraphics[width=\textwidth]{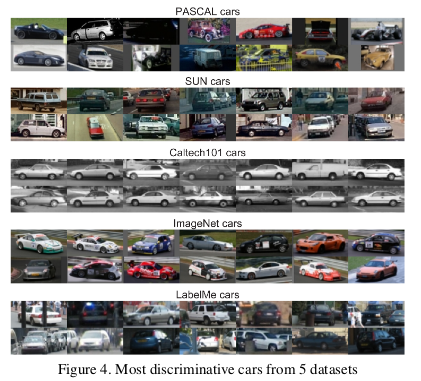}

But there are biases in the data that make each dataset distinctive. For example, there are many race cars in the ImageNet dataset, which cannot be said to represent the "platonic" concept of a car in general.

\includegraphics[width=\textwidth]{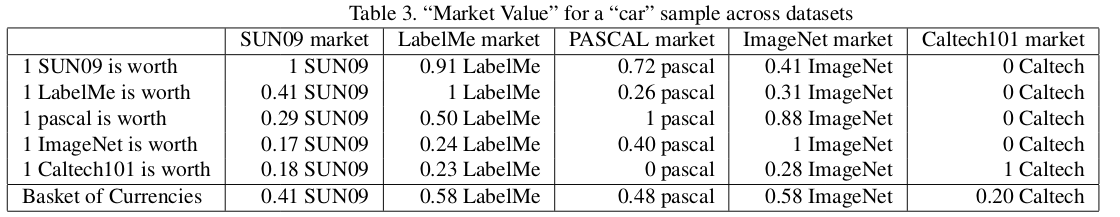}

The authors further judged the value of a dataset by measuring how well a classifier trained on it performs on other datasets. By this metric, LabelMe and ImageNet are the least biased datasets, scoring $ 0.58 $ in a "basket of currencies." The values are all less than one, which means that training on a different dataset always results in lower test performance. In an ideal world without dataset bias, some of these values should be above one.

The authors pessimistically concluded:
\begin{quote}
So, what is the value of current datasets when used to train algorithms that will be deployed in the real world? The answer that emerges can be summarized as: “better than nothing, but not by much
\end{quote}

\section*{Myth 3: Machine Learning researchers do not use the test set for validation}
\label{sec:myth-3}
In Machine Learning 101, we are taught to split a dataset into training, validation, and test sets. The performance of a model trained on the training set and evaluated on the validation set helps the machine learning practitioner tune his model to maximize its performance in real world usage. The test set should be held out until the practitioner is done with the tuning so as to provide an unbiased estimate of the model's actual performance in real world usage. If the practitioner "cheats" by using the test set in the training or validation process, he runs the risk of overfitting his model to biases inherent in the dataset that do not generalize beyond the dataset.

In the hyper-competitive world of machine learning research, new algorithms and models are often evaluated using their performance on the test set. Thus, there is little reason for researchers to write or submit papers that propose methods with inferior test performance. This effectively means that the machine learning research community, as a whole, is using the test set for validation.

What is the impact of this "cheating"?

\includegraphics[width=\textwidth]{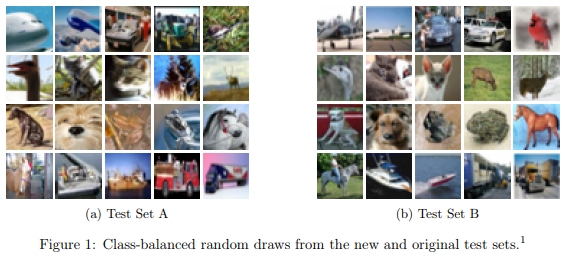}

The authors of \citet{recht2018cifar} investigated this by creating a new test set for CIFAR-10. They did this by parsing images from the Tiny Images repository, as was done in the original dataset collection process.

They chose CIFAR-10 because it is one of the most widely used datasets in machine learning, being the second most popular dataset in NeurIPS 2017 (after MNIST). The dataset creation process for CIFAR-10 is also well-documented and transparent, with the large Tiny Images repository having sufficiently fine-grained labels that make it possible to replicate a new test set while minimizing distributional shift.

\includegraphics[width=\textwidth]{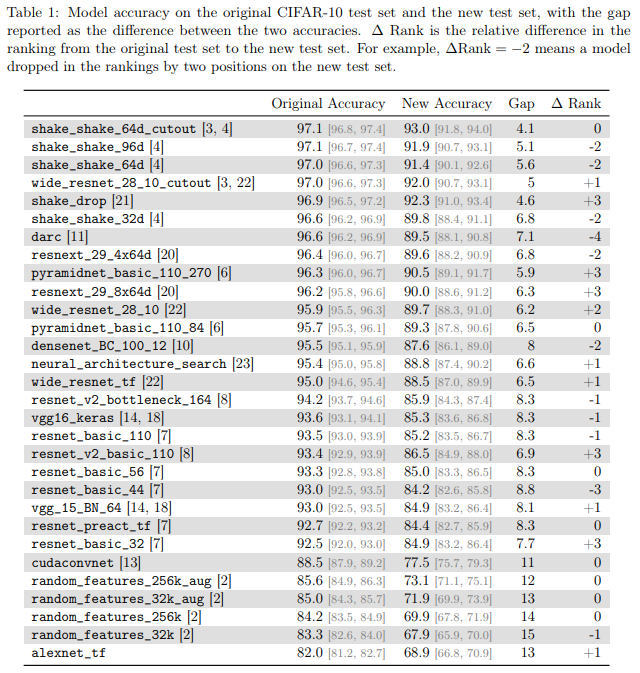}

They found that across a wide range of different neural network models, there was a significant drop in accuracy ($ 4\% - 15\% $) from the old test set to the new test set. However, the relative ranking of each model's performance remained fairly stable.

\includegraphics[width=\textwidth]{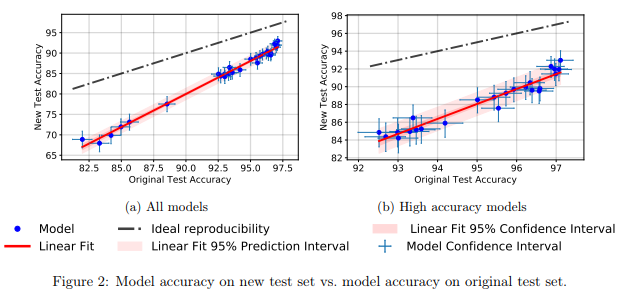}

In general, the higher performing models experienced a smaller drop in accuracy compared to the lower performing models. This is heartening, because it indicates that the loss in generalization caused by the "cheating," at least in the case of CIFAR-10, becomes more muted as the research community invents better machine learning models and methods.

\section*{Myth 4: Every datapoint is used in training a neural network}
\label{sec:myth-4}
Conventional wisdom says that \href{https://www.economist.com/leaders/2017/05/06/the-worlds-most-valuable-resource-is-no-longer-oil-but-data}{data is the new oil} and the more data we have, the better we can train our sample-inefficient and overparametrized deep learning models.

In \citet{toneva2018empirical}, the authors demonstrate significant redundancy in several common small image datasets. Shockingly, $ 30\% $ of the datapoints in CIFAR-10 can be removed, without changing test accuracy by much.

\includegraphics[width=\textwidth]{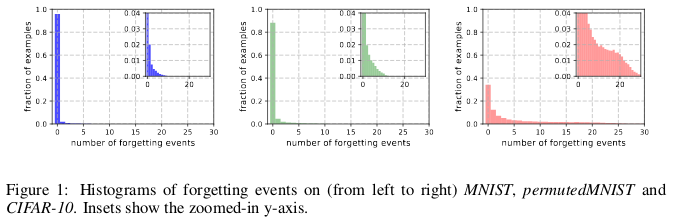}

A forgetting event happens when the neural network makes a misclassification at time $t+1$, having already made an accurate classification at time $t$, where we consider the flow of time to be the number of SGD updates made to the network. To make the tracking of forgetting events tractable, the authors run their neural network over only the examples in the mini-batch every time an SGD update is made, rather than over every single example in the dataset. Examples that do not undergo a forgetting event are called \emph{unforgettable} examples.

They find that $91.7\%$ of MNIST, $75.3\%$ of permutedMNIST, $31.3\%$ of CIFAR-10, and $7.62\%$ of CIFAR-100 comprise of unforgettable examples. This makes intuitive sense, since an increase in the diversity and complexity of an image dataset should cause the neural network to forget more examples.

\includegraphics[width=\textwidth]{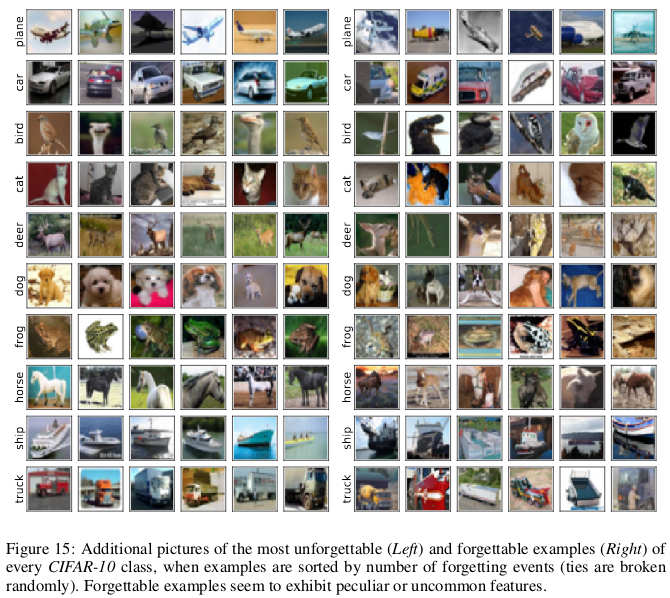}

Forgettable examples seem to display more uncommon and peculiar features than unforgettable examples. The authors liken them to support vectors in SVM, because they seem to demarcate the contours of the decision boundary.

\includegraphics[width=\textwidth]{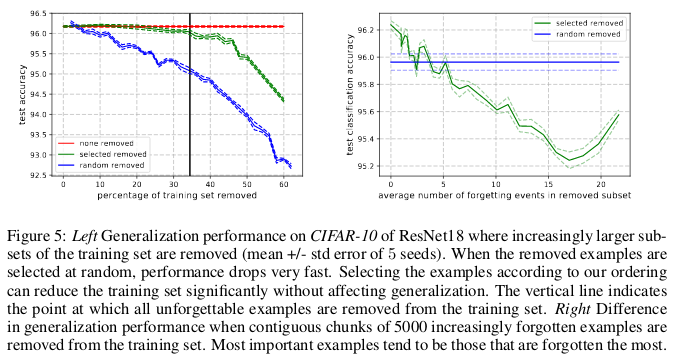}

Unforgettable examples, by contrast, encode mostly redundant information. If we sort the examples by their unforgettability, we can compress the dataset by removing the most unforgettable examples.

On CIFAR-10, $30\%$ of the dataset can be removed without affecting test accuracy, while a $35\%$ removal causes a trivial $0.2\%$ dip in test accuracy. If this $30\%$ was selected by random instead of chosen by unforgettability, then its removal will result in a significant loss of around $1\%$ in test accuracy.

\includegraphics[width=\textwidth]{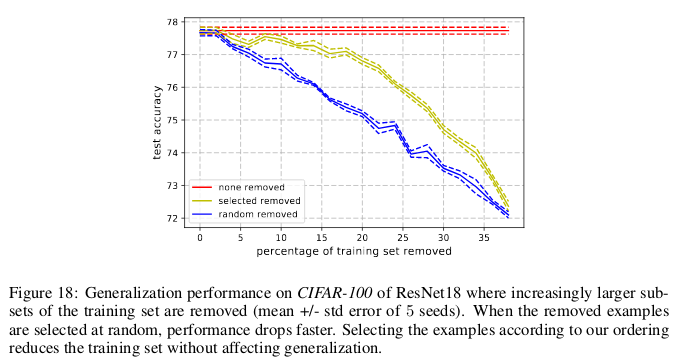}

Similarly, on CIFAR-100, $8\%$ of the dataset can be removed without affecting test accuracy.

These findings show that there is significant data redundancy in neural network training, much like in SVM training where the non-support vectors can be taken away without affecting the decisions of the model.

\subsection*{Implication:}
If we can determine which examples are unforgettable before the start of training, then we can save space by removing those examples and save time by not training the neural network on them.

\section*{Myth 5: We need (batch) normalization to train very deep residual networks}
\label{sec:myth-5}
It had been believed for a very long time that "training a deep network to directly optimize only the supervised objective of interest (for example the log probability of correct classification) by gradient descent, starting from random initialized parameters, does not work very well." \citep{vincent2010stacked}

Since then, a host of clever random initialization methods, activation functions, optimization techniques, and other architectural innovations like residual connections \citep{he2016deep}, has made it easier to train deep neural networks with gradient descent.

But the real breakthrough came from the introduction of batch normalization \citep{ioffe2015batch} (and other subsequent normalization techniques), which constrained the size of activations at every layer of a deep network to mitigate the vanishing and exploding gradients problem.

In a recent paper \citet{zhang2019fixup}, it was shown remarkably that it is actually possible to train a $10,000$-layer deep network using vanilla SGD, without resorting to any normalization.

\includegraphics[width=\textwidth]{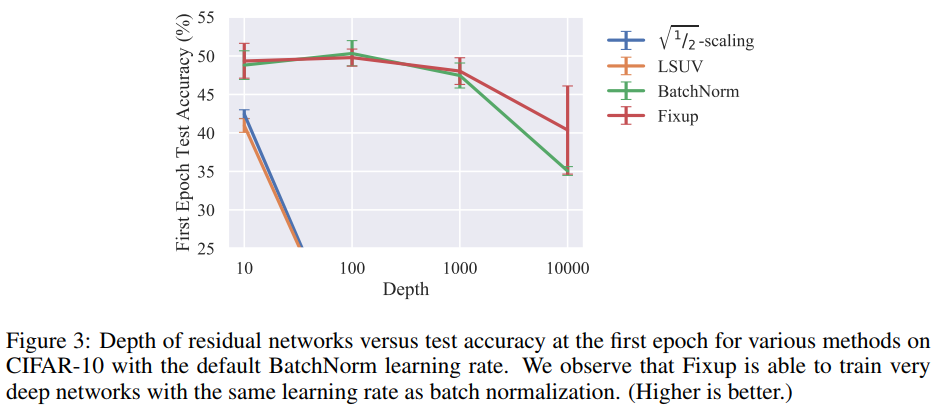}

The authors compared training a residual network at varying depths for one epoch on CIFAR-10, and found that while standard initialization methods failed for $100$ layers, both Fixup and batch normalization succeeded for $10,000$ layers.

\includegraphics[width=\textwidth]{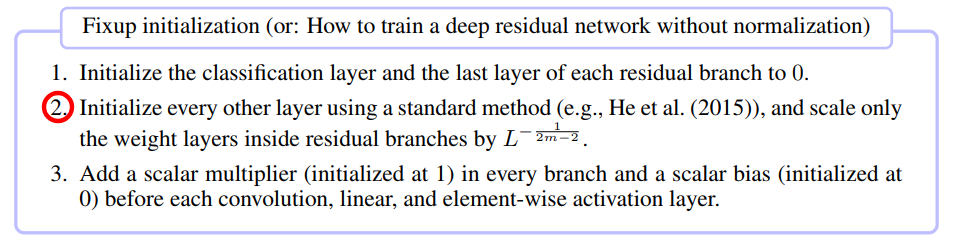}

They did a theoretical analysis to show that "the gradient norm of certain layers is in expectation lower bounded by a quantity that increases indefinitely with the network depth," i.e. the exploding gradients problem.

To prevent this, the key idea in Fixup is to scale the weights in the $m$ layers for each of $L$ residual branches by a factor that depends on $m$ and $L$.

\includegraphics[width=\textwidth]{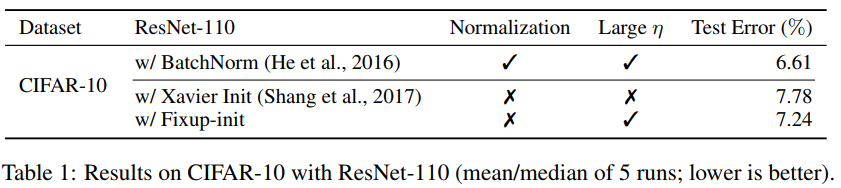}

Fixup enabled the training of a deep residual network with $110$ layers on CIFAR-10 with a large learning rate, at comparable test performance with the same network architecture that had batch normalization.

\includegraphics[width=\textwidth]{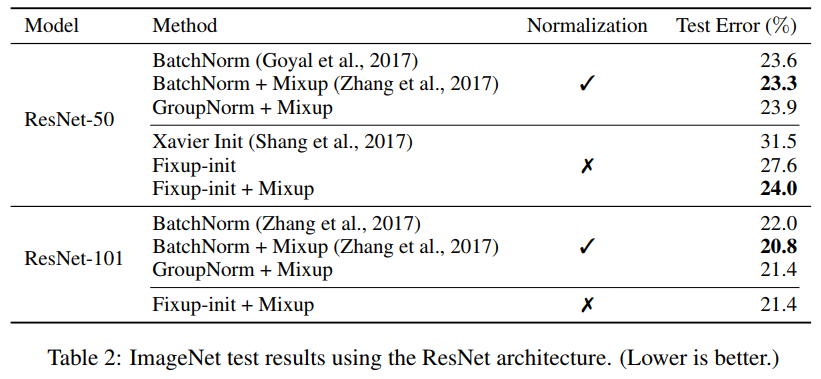}
\includegraphics[width=\textwidth]{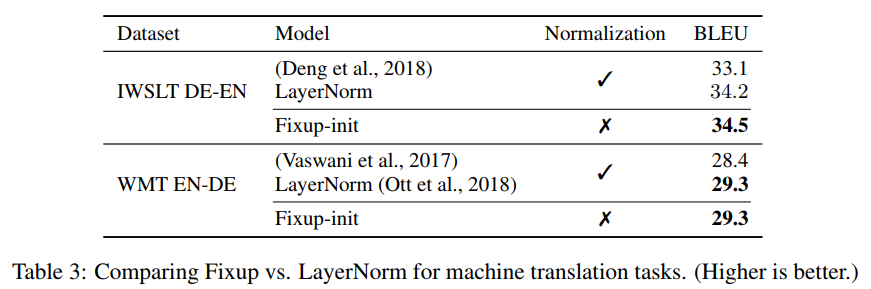}

The authors also further showed comparable test results using a Fixup-ed network without any normalization on the ImageNet dataset and English-German machine translation tasks.

\section*{Myth 6: Attention $>$ Convolution}
\label{sec:myth-6}
There is an idea gaining currency in the machine learning community that attention mechanisms are a superior alternative to convolutions now \citep{vaswani2017attention}. Importantly, \citet{vaswani2017attention} noted that "the computational cost of a separable convolution is equal to the combination of a self-attention layer and a point-wise feed-forward layer."

Even state-of-the-art GANS find self-attention superior to standard convolutions in its ability to model long-range, multi-scale dependencies \citep{zhang2018self}.

The authors of \citet{wu2019pay} question the parameter efficiency and efficacy of self-attention in modelling long-range dependencies, and propose new variants of convolutions, partially inspired by self-attention, that are more parameter-efficient.

\includegraphics[width=\textwidth]{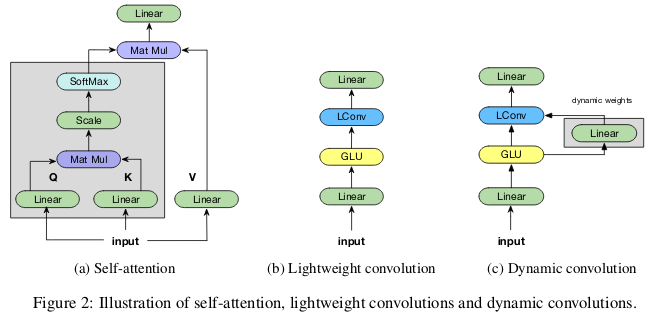}

\emph{Lightweight} convolutions are depthwise-separable, softmax-normalized across the temporal dimension, shares weights across the channel dimension, and re-uses the same weights at every time step (like RNNs). \emph{Dynamic} convolutions are lightweight convolutions that use different weights at every time step.

These tricks make lightweight and dynamic convolutions several orders of magnitude more efficient that standard non-separable convolutions.

\includegraphics[width=\textwidth]{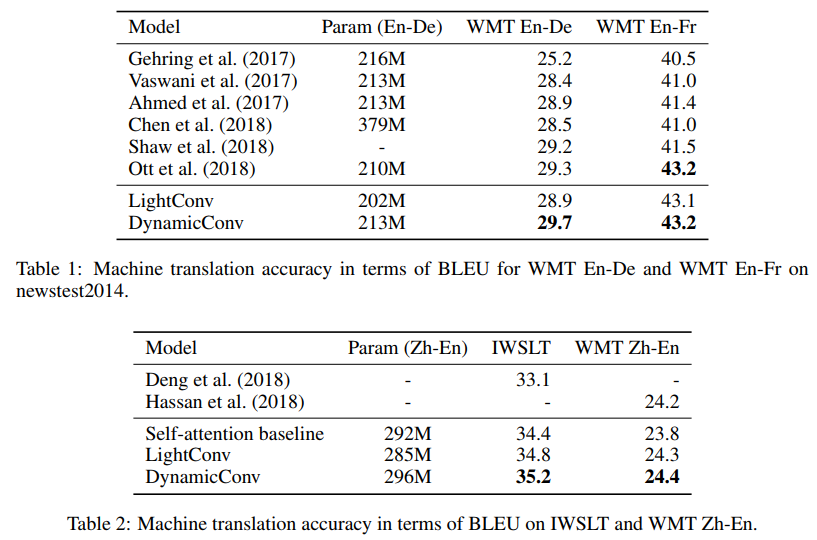}
\includegraphics[width=\textwidth]{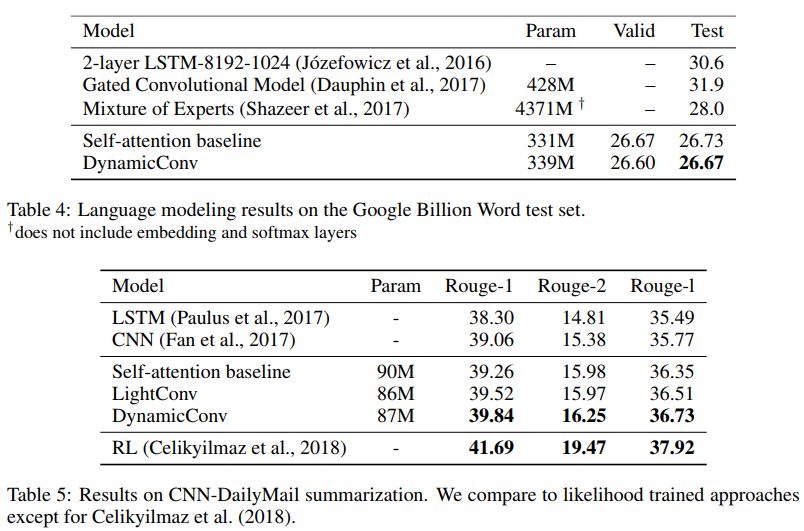}

The authors show that these new convolutions match or exceed the self-attention baselines in machine translation, language modelling, and abstractive summarization tasks while using comparable or less number of parameters.

\section*{Myth 7: Saliency maps are robust ways to interpret neural networks}
\label{sec:myth-7}
While neural networks are commonly believed to be black boxes, there have been many, many attempts made to interpret them. Saliency maps, or other similar methods that assign importance scores to features or training examples, are the most popular form of interpretation.

It is tempting to be able to conclude that the reason why a given image is classified a certain way is due to particular parts of the image that are salient to the neural network's decision in making the classification. There are several ways to compute this saliency map, often making use of a neural network's activations on a given image and the gradients that flow through the network.

In \citet{ghorbani2017interpretation}, the authors show that they can introduce an imperceptible perturbation to a given image to distort its saliency map.

\includegraphics[width=\textwidth]{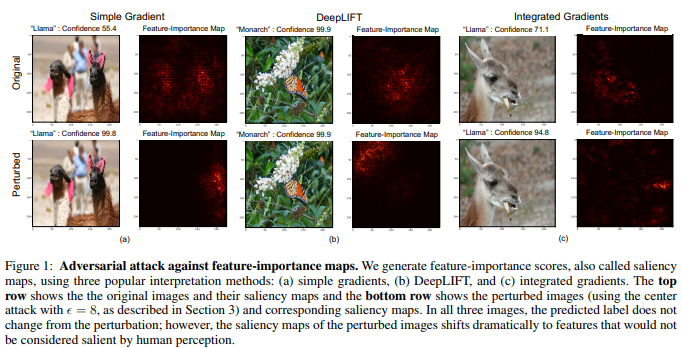}

A monarch butterfly is thus classified as a monarch butterfly, not on account of the patterns on its wings, but because of some unimportant green leaves in the background.

\includegraphics[width=\textwidth]{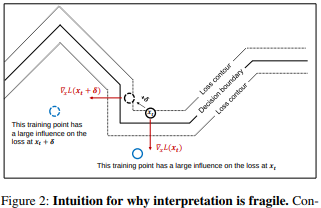}

High-dimensional images often lie close to the decision boundaries constructed by deep neural networks, hence their susceptibility to adversarial attacks. While adversarial attacks shift images past a decision boundary, adversarial interpretation attacks shift them along the contour of the decision boundary, while still remaining within the same decision territory.

\includegraphics[width=\textwidth]{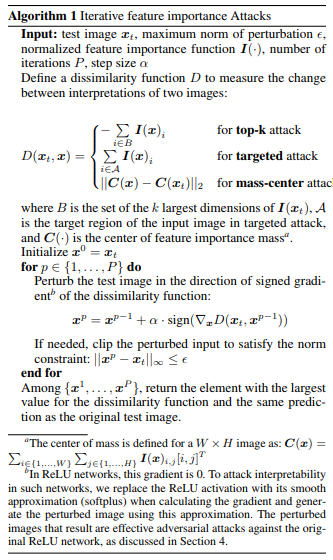}

The basic method employed by the authors to do this is a modification of \citet{goodfellow2014explaining}'s fast gradient sign method, which was one of the first efficient adversarial attacks introduced. This suggests that other more recent and sophisticated adversarial attacks can also be used to attack neural network interpretations.

\includegraphics[width=\textwidth]{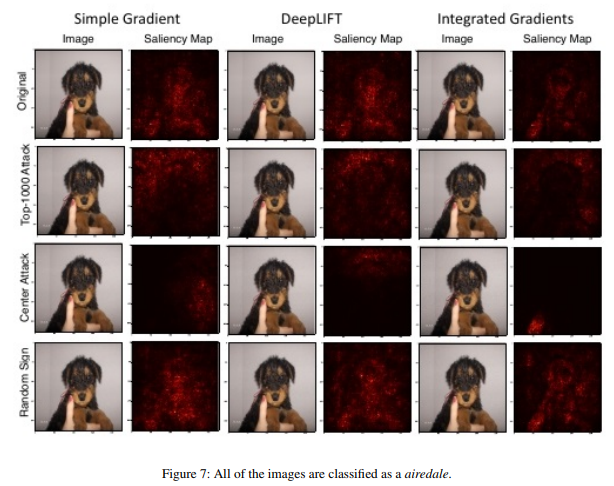}
\includegraphics[width=\textwidth]{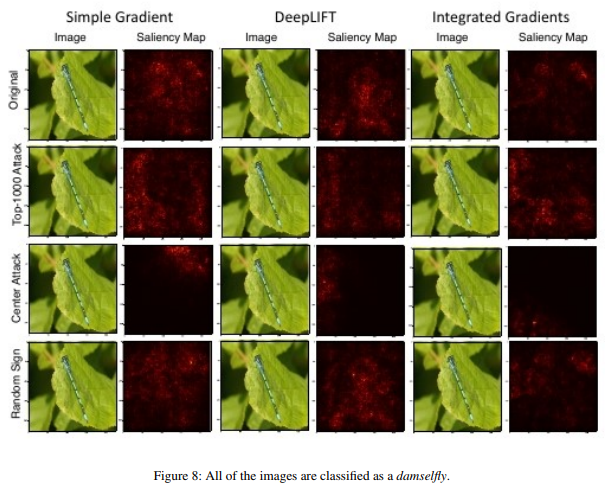}
\includegraphics[width=\textwidth]{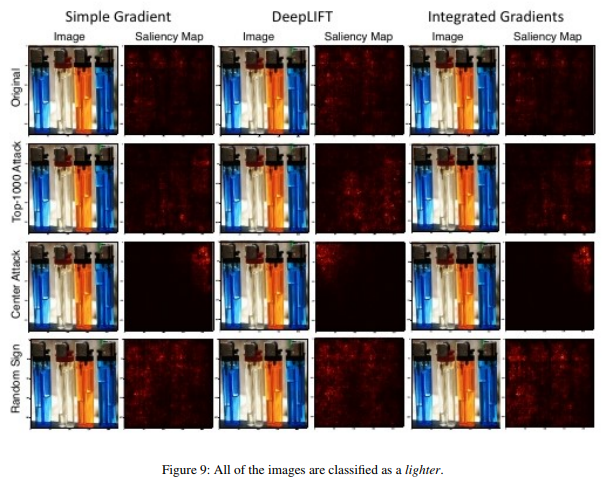}

\subsection*{Implication:}
As deep learning becomes more and more ubiquitous in high stakes applications like medical imaging, it is important to be careful of how we interpret decisions made by neural networks. For example, while it would be nice to have a CNN identify a spot on an MRI image as a malignant cancer-causing tumor, these results should not be trusted if they are based on fragile interpretation methods.

\bibliography{bib}

\end{document}